\documentclass[conference]{IEEEtran}
\IEEEoverridecommandlockouts
\usepackage{cite}
\usepackage{amsmath,amssymb,amsfonts}
\usepackage{graphicx}
\usepackage{textcomp}

\usepackage{graphics} 
\usepackage[linesnumbered,ruled]{algorithm2e}
\graphicspath{{img/}}
\usepackage{epsfig}
\usepackage[english]{babel}
\usepackage{changepage}
\usepackage[graphicx]{realboxes}
\usepackage{adjustbox}
\usepackage{footnote}
\usepackage{pifont}
\usepackage{cite}
\usepackage{tablefootnote} 
\usepackage[makeroom]{cancel}
\usepackage[table]{xcolor} %
\usepackage{multirow}
\usepackage{booktabs}
\usepackage{subfigure}
\usepackage{caption}
\usepackage{epstopdf}
\usepackage{bm}

\usepackage{algpseudocode,algorithm} 
\usepackage{color}
\usepackage{longtable}
\usepackage{bm,nicefrac}
\usepackage{mathrsfs}
\usepackage{hhline}
\usepackage{pdflscape}
\usepackage{balance}
\usepackage{ltablex}
\usepackage{hyperref}
\usepackage{threeparttablex}
\usepackage{enumitem}
\usepackage{units}
\usepackage{mathtools}
\usepackage{dsfont}
\usepackage{comment}
\usepackage[colorinlistoftodos]{todonotes}

\usepackage[normalem]{ulem}
\usepackage{tikz}

\def\BibTeX{{\rm B\kern-.05em{\sc i\kern-.025em b}\kern-.08em
    T\kern-.1667em\lower.7ex\hbox{E}\kern-.125emX}}
\begin{document}




\title{Out-of-distribution Object Detection\\ through Bayesian Uncertainty Estimation} 



%

\author{Tianhao Zhang$^1$, Shenglin Wang$^1$, Nidhal Bouaynaya$^2$, Radu Calinescu$^3$ and Lyudmila Mihaylova$^1$\\
\textit{$^1$ Automatic Control and Systems Engineering, 
The University of Sheffield, UK}\\
\textit{$^2$ Department of Electrical and Computer
Engineering, Rowan University, USA}\\
\textit{$^3$ Department of Computer Science, University of York, UK}\\
Email: \{tzhang81@sheffield.ac.uk, swang119@sheffield.ac.uk, bouaynaya@rowan.edu,\\ radu.calinescu@york.ac.uk, l.s.mihaylova@sheffield.ac.uk\} }

\maketitle
\begin{abstract}
The superior performance of object detectors is often established under the condition that the test samples are in the same distribution as the training data. However, in many practical applications, out-of-distribution (OOD) instances are inevitable and usually lead to uncertainty in the results. In this paper, we propose a novel, intuitive, and scalable probabilistic object detection method for OOD detection. Unlike other uncertainty-modeling methods that either require huge computational costs to infer the weight distributions or rely on model training through synthetic outlier data, our method is able to distinguish between in-distribution (ID) data and OOD data via weight parameter sampling from proposed Gaussian distributions based on pre-trained networks. We demonstrate that our Bayesian object detector can achieve satisfactory OOD identification performance by reducing the FPR95 score by up to 8.19\% and increasing the AUROC score by up to 13.94\% when trained on BDD100k and VOC datasets as the ID datasets and evaluated on COCO2017 dataset as the OOD dataset.

\end{abstract}

\begin{IEEEkeywords} 
Out-of-distribution detection, Uncertainty estimation, object detection, deep learning, image classification
\end{IEEEkeywords}


\begin{figure}[tpb]
\centering
\includegraphics[width=0.75\linewidth, height=0.9\linewidth]{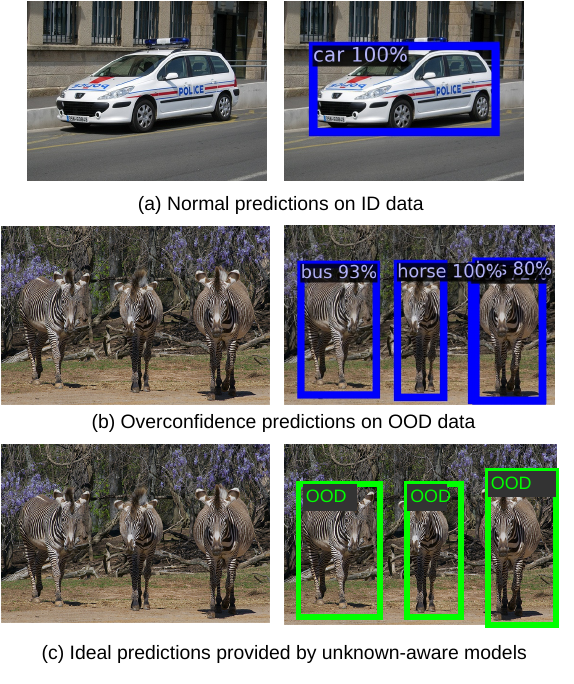}
\caption{\small (a) An example prediction of PASCAL VOC dataset~\cite{DBLP:journals/ijcv/EveringhamGWWZ10} provided by a standard Faster-RCNN~\cite{DBLP:journals/pami/RenHG017} model trained on the same dataset. (b) The same Faster-RCNN model produces overconfident predictions on MS-COCO dataset~\cite{lin2014microsoft}, which is an OOD dataset in this case. (c) The ``ideal'' predictions provided by the uncertainty-aware object detection model can recognize OOD data.}\label{fig:exp}
\end{figure}

\section{Introduction}\label{sec:introduction}
Out-of-distribution (OOD) detection is a machine learning technique that aims to detect test samples drawn from a distribution that differs from the distribution of the training data, with the definition of distribution to be well-defined according to the application in the target. The goal of OOD detection is to differentiate between inputs that are likely to be part of the distribution the model was trained on, and inputs that are not~\cite{yang2021generalized}. The vast majority of modern deep neural networks~\cite{sze2017efficient} are deterministic models that provide high-confidence predictions for inputs that are not seen during training, leading to poor generalization and unreliable results, as shown for example in Figure~\ref{fig:exp}. OOD detection is particularly important for safety-related scenarios. For instance, an ideal uncertainty-aware deep object detector built for self-driving cars should be able to recognize the in-distribution (ID) target, such as people and cars, and produce a low-confidence prediction for the OOD objects, such as an animal. 

Exploring the uncertainties of the deep learning algorithms can help perform OOD detection~\cite{yang2021generalized}. In general, there are two types of uncertainties: \textit{aleatoric uncertainty}, due to the inherent and irreducible uncertainty of the input data, and \textit{epistemic uncertainty} due to various deep learning model frameworks~\cite{kendall2017uncertainties}. A Bayesian Neural Networks (BNNs) is a type of neural network that incorporates Bayesian methods, such as probability distributions over model parameters, to represent these two types of uncertainties in the model's predictions~\cite{neal1995bayesian}. BNNs can be implemented for OOD detection by comparing the uncertainties between the model's predictions on a given input and the model's predictions on a set of known ID data. If the uncertainty in the input is high, the input is likely OOD. Also, Bayesian methods can be used to estimate the likelihood of the data under the model. If this likelihood is low, the input is more likely to be OOD data.  
However, due to the high dimensionality and multi-modality of modern deep neural networks with millions of weight parameters, Bayesian approaches have largely been intractable for state-of-the-art deep learning object detectors~\cite{10.1007/978-3-031-20080-9_37}. State-of-the-art deterministic CNNs have achieved huge success on object detection tasks~\cite{wang2022internimage, DBLP:journals/corr/abs-2012-07177, dagli2021cppe5}. A new inquiry is whether it is possible to create a Bayesian object detector that can maintain its outstanding performance on the ID dataset and have the ability to model uncertainty when it encounters the OOD dataset concurrently.


In order to address this problem, we propose a novel and scalable uncertainty-aware Bayesian deep learning method for OOD object detection. We exploit the information contained in the deterministic deep neural network layers when the model is trained in the ID dataset to efficiently approximate the posterior distribution over the weights. Our Bayesian approach focuses on performing Bayesian inference by sampling the predictions from the proposed approximation Gaussian distributions of neural weights. Our Bayesian inference method transforms the deterministic deep neural network layers into probabilistic Bayesian layers with no additional training cost. It makes it possible to maximize the uncertainty quantification abilities of BNNs on large computer vision tasks via BNNs.




In particular, our main contributions are: 
\begin{itemize}
 \item In this work, we propose a scalable and flexible approximate Bayesian inference technique for OOD detection for deep learning object detectors. Our method offers the ability of uncertainty estimation to distinguish the OOD data from the ID data while preserving their outstanding performance on the ID task for large deep learning models. Our framework provides a flexible uncertainty estimation approach by choosing different layers transformed into Bayesian layers during the model inference stage. Different levels of uncertainties can be chosen to be reserved and estimated.  
 
 
 \item Our proposed Bayesian inference technique has been demonstrated effective on OOD detection tasks by comprehensively evaluated on typical OOD detection benchmarks. We show in Section~\ref{sec:results} that compared to other Bayesian methods, our method can reduce the FPR95 score by up to 8.19\% and increase the AUROC score by up to 13.94\% on BDD and PASCAL VOC as ID datasets and COCO2017 as the OOD dataset.
\end{itemize}

The rest of the paper is organized in the following way. Section~\ref{sec:bg} provides the background of the OOD detection and Bayesian deep learning methods. In Section~\ref{sec:method}, the proposed Bayesian inference framework is introduced in detail. \ref{sec:results} shows the comprehensive experimental results and analysis on OOD benchmarks on both object detection and image classification tasks. The conclusions are presented in Section~\ref{sec:conclusion}.
\begin{figure*}[t]
	\centering
        \includegraphics[width= 0.9\textwidth]{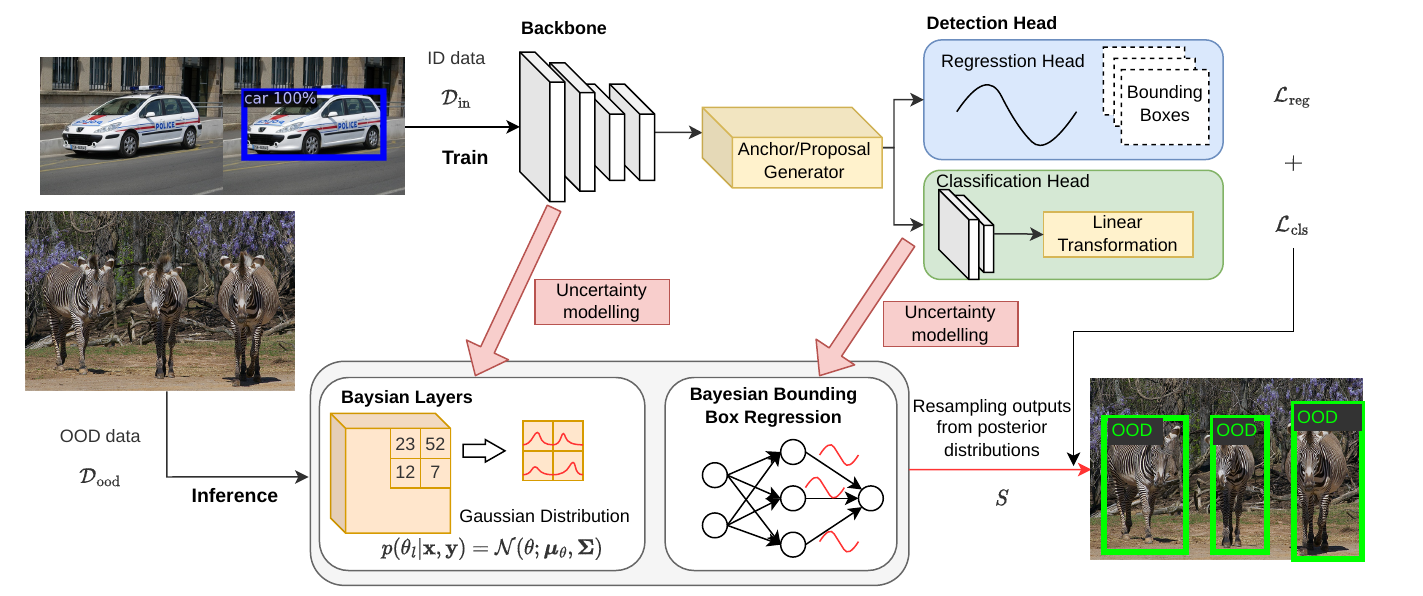}
	\caption{The framework of our proposed Bayesian inference method. The base model is first trained on the ID dataset with the joint object detection loss ($\mathcal{L}_{\mathrm{reg}},\mathcal{L}_{\mathrm{cls}}$). During the inference stage, we assume the distributions of weight parameters from the chosen layers as class-conditional Gaussians, and sample the results from the low-likelihood region given by the OOD data. The uncertainty estimation produced by the Bayesian layers coming out of the backbone, the classification head or the regression head provides a high uncertainty score $S$ for the OOD data. Combined with the OOD scoring rule with the uncertainty modelling by the Bayesian layers, the OOD data can be separated from the ID data.} 
        \label{fig:method}
\end{figure*}

\section{Related Work}\label{sec:bg}
\subsection{Out-of-distribution Detection}
The goal of out-of-distribution detection, or OOD detection, is to identify input data coming out of a different distribution from the training distribution. OOD detection is an important technique used to help neural networks determine their capability boundary~\cite{Diffusion}. More specifically, given a labeled training dataset of $N$ data pairs as $\mathcal{D} = \{\mathbf{x}_n, \mathbf{y}_n\}_{n=1}^{N}$, where $\mathbf{x}_n$ is an input data sample in the domain $\mathcal{X}$, and $\mathbf{y}_n$ its corresponding target value in the domain $\mathcal{Y}$. The goal of OOD detection is to build a detector $f$ that $f(x_{1},...,x_{n})=1, \forall _{i}, p(x_{i})\geq \delta$ and $f(x_{1},...,x_{n})=0, \forall _{i}, p(x_{i})\leq  \delta$ where $\delta$ is the capability boundary. 

\textbf{OOD detection for classification} can be broadly categorized into two main approaches: generative-based methods~\cite{DBLP:journals/corr/abs-1810-01392, ren2019likelihood, serra2019input, xiao2020likelihood} and reconstruction-based methods\cite{lyudchik2016outlier,pimentel2014review,denouden2018improving,zong2018deep}. A conceptually appealing approach to OOD detection involves fitting a generative model  to a data distribution $p(\mathbf{x}; \theta)$ and evaluating the likelihood of unseen samples under this model. The assumption is that OOD samples will be assigned a lower likelihood than in-distribution samples and can be identified using a simple threshold on this value~\cite{bishop1994novelty}. Some of the state-of-the-art algorithms are: ensembling~\cite{DBLP:conf/nips/Lakshminarayanan17},  ODIN~\cite{liang2018enhancing}, energy score~\cite{liu2020energy}, Mahalanobis distance~\cite{lee2018simple}, Gram matrices
based score~\cite{DBLP:conf/icml/SastryO20}, and GradNorm score~\cite{huang2021importance}.


\textbf{OOD detection for object detection} is currently underexplored~\cite{DBLP:journals/corr/abs-2103-02603, 9922026, 9933263}. Energy-based~\cite{liu2020energy} has been proved with both mathematical insights and empirical evidence that the energy score is superior to both a softmax-based score and generative-based methods for OOD detection. Based on directly predicting the conditional target density of an energy-based model, a novel framework for OOD detection by adaptively synthesizing virtual outliers that can meaningfully regularize the model's decision boundary during training known as VOS~\cite{DBLP:journals/corr/abs-2103-02603} has been proved effective on OOD detection by modeling the aleatoric uncertainty. As a comparison, our method combines the uncertainty estimation for the classification and localization regression with a novel OOD scoring rule.

\subsection{Bayesian Methods for Probabilistic Object Detection}
Different from deterministic neural networks, BNNs aim to provide a natural interpretation of uncertainty estimation in deep learning, by inferring distributions over a network's weights $\theta \in \mathcal{W}$ based on the training dataset $\mathcal{D}$. 

BNNs produce the predictive distribution $p(\mathbf{y}|\mathbf{x},\mathcal{D})$ by integrating over all values of network weights:
\begin{equation} \label{eq:bnn}
p(\mathbf{y}|\mathbf{x}, \mathcal{D}) = \int p(\mathbf{y}|\mathbf{x}, \theta) p(\theta|\mathcal{D}) \mathrm{d} \theta,
\end{equation}
where $p(\mathbf{y}|\mathbf{x}, \theta)$ represents the predicted distribution, and $p(\theta|\mathcal{D})$ represents the weight posterior distribution over the dataset~\cite{9525313}.

BNNs can be used for OOD detection by comparing the uncertainty of the model's predictions on a given input to the uncertainty of the model's predictions on a set of known in-distribution data. In order to calculate the approximate solutions of the posterior distribution, various techniques have been proposed~\cite{neal2012bayesian, chen2014stochastic, blundell2015weight,graves2011practical,depeweg2018decomposition,neal2012bayesian,welling2011bayesian,chen2014stochastic,chen2020statistical,mandt2017stochastic,maddox2019simple,ritter2018scalable}. However, due to the high dimensionality and multi-modality of modern deep neural networks with millions of weight parameters, the posterior distribution is intractable for most of the state-of-the-art object detectors.

\textbf{Markov Chain Monte Carlo (MCMC)} methods are well-recognised methods for inference, including those combined with neural networks, for instance, the Hamiltonian Monte Carlo (HMC)~\cite{neal2012bayesian}. However, the biggest disadvantage is that requiring full gradients makes HMC computationally intractable for modern neural networks. Later the HMC framework has been extended into stochastic gradient HMC (SGHMC)~\cite{chen2014stochastic}, which allows stochastic gradients to be used in Bayesian inference. As an alternative, stochastic gradient Langevin dynamics (SGLD)~\cite{welling2011bayesian} employs first-order Langevin dynamics in the stochastic gradient process. 

\textbf{Monte-Carlo Dropout}
aims to reduce the computational cost for approximating the true posterior distributions~\cite{Gal2016Uncertainty}. It is a Bayesian inference method implemented based on the regularization techniques with Stochastic Gradient Descent (SGD) during the training stage, which links dropout-based neural network training to Variational Inference (VI) in BNNs. This method is later extended~\cite{gal2017concrete} by proving that the optimal dropout rate can be found by treating the dropout probability as a hyperparameter. During test time, samples from the approximate posterior distribution are generated by performing inference multiple times with dropout enabled: $p(\mathbf{y}|\mathbf{x}, \mathcal{D}) \approx \frac{1}{T} \sum_{t=1}^{T} p(\mathbf{y}|\mathbf{x}, \theta_t)$. However, recent researchers are questioning whether MC Dropout is a Bayesian method~\cite{DBLP:journals/corr/abs-2110-04286}. The implementation of MC Dropout fundamentally depends on the dropout layers being used as the regularization method during the training stage. Since 2015, Batch Normalization (BN)~\cite{ioffe2015batch} has been applied to normalize the activations of a layer in a neural network by adjusting and scaling the activations to reduce internal covariate shifts. Due to the problem of variance shift problem when using BN and Dropout at the same time~\cite{8953671}, BN has been proven as a more effective normalization method, which leads to Dropout being barely used in modern deep neural networks. This makes it difficult to use MC Dropout as a Bayesian inference method in advanced deep-learning models.

\textbf{Direct Modeling Methods} 
refer to those methods which assume certain probability distributions over the weight parameters in the neural networks, and then directly predict parameters for such distributions through output layers~\cite{9525313}. Some methods aim at modifying the network’s output layers and the loss
function to model the uncertainty through a single forward pass. A softmax function with a Gaussian distribution is used to replace the standard softmax function with the cross-entropy loss~\cite{kendall2017uncertainties} for classification uncertainty quantification. For object detection tasks, some methods focused on modeling the uncertainty coming out of the bounding box regression~\cite{harakeh2019bayesod, he2019bounding}. Furthermore, several works propose to estimate higher-order conjugate priors in addition to directly predicting the output probability distributions~\cite{amini2019deep}. Stochastic Gradient Descent (SGD) Based Approximations~\cite{DBLP:journals/corr/abs-1708-02182} use the iterates of averaged SGD as an MCMC sampler. Stochastic Weight Average Gaussian (SWAG)~\cite{maddox2019simple} for Bayesian model averaging and uncertainty estimation proves that the posterior distribution over neural network parameters is close to Gaussian in the subspace spanned by the trajectory of SGD.

\section{Method}\label{sec:method}

Our proposed Bayesian object detection inference framework is illustrated in Figure~\ref{fig:method}. One significant advantage of our method is that we can select which layers of the model to sample. Different levels of uncertainties can be represented by choosing the different numbers of Bayesian layers. The experimental results of choosing different layers to perform OOD detection are shown in section~\ref{sec:results}. 

\subsection{Bayesian Neural Networks for Object Detection}
\label{sec:bnn}
Followed by the definition of the data in section~\ref{sec:bg}, the goal of OOD detection for object detection is to provide predictions based on the training data $\mathcal{D} = \{\mathbf{x}_i, \mathbf{b}_i, \mathbf{y}_i\}_{i=1}^{N}$, where $\mathbf{x} \in \mathcal{X}$ denotes the input image, $\mathbf{b} \in \mathbb{R}^4$ represents the bounding box coordinates associated with object instances in the image, and $\mathbf{y} \in \mathcal{Y}$ is semantic labels to its corresponding target for $K$ classes classification. In stead of forming the parameters  through maximum a-posterior (MAP) optimization $\hat{\boldsymbol\theta}_{\text{MAP}} = \text{argmax}_{\theta} p({\theta} | \mathcal{D})$, BNNs aim to produce the predictive classification distribution $p_{\boldsymbol\theta}(\hat{\mathbf{y}}|\hat{\mathbf{x}})$ and the bounding box regression distribution $p_{\theta}(\hat{\mathbf{b}} |\hat{\mathbf{y}},\hat{\mathbf{x}})$ through a Monte Carlo sampling procedure: $p(\hat{\mathbf{y}} | \mathcal{D}, \hat{\mathbf{x}}) \approx \frac{1}{T} \sum^T_{t=1} p(\hat{\mathbf{y}} | \theta_l, \hat{\mathbf{x}}) \,, \quad {\boldsymbol\theta}_l \sim p(\theta | \mathcal{D})$, where the $\hat{\mathbf{y}}$ denotes the output predictions of the test input data $\hat{\mathbf{x}}$, and $\hat{\mathbf{b}}$. 

While the straightforward idea is to infer the posterior distribution $p(\mathbf{y}|\mathbf{x}, {\theta})$ based on the training dataset $\mathcal{D}$ during the training stage. Variational Inference (VI) implies fitting a Gaussian variational posterior approximation over the weights of neural networks, and then approximating the prior distribution through Kullback–Leibler (KL) divergence~\cite{graves2011practical}. VI is later generalized through the reparameterization trick for DNNs~\cite{blundell2015weight, kingma2015variational}. Same as other Bayesian approximation methods introduced in Section~\ref{sec:bg}, VI methods are empirically hard to be deployed on deep neural networks with numerous parameters. 
\subsection{Weights Sampling from the Bayesian Layers}
Instead of training to approximate the posterior distribution $p(\mathbf{y}|\mathbf{x}, \theta)$, our key idea is that, given a pre-trained model, we assume the uncertainty representation of the weight parameters of different Bayesian layers from class-conditional multivariate Gaussian distributions:
\begin{equation}
    p(\theta_{l}| \mathbf{x}, \mathbf{y})=\mathcal{N}(\theta ; {\boldsymbol\mu}_\theta , \mathbf{\boldsymbol\Sigma})
    \label{eq:Guassian}
\end{equation}
where ${\boldsymbol\mu}_\theta$ is the Gaussian mean, $\*{\boldsymbol\Sigma}$ is the related covariance matrix, and $\theta_{l}$ is weight parameters corresponding to different layers in the neural networks. The ability to model epistemic uncertainty can be achieved by sampling the weights from the proposed prior distribution.

To formulate the parameters of the proposed class-conditional Gaussian distributions, we propose that the sample class means $\hat{\bm\mu}_\theta$ are chosen as the values of the weight parameters ${\theta}_{pre}$ from the pre-trained models. The true posterior weight distribution $p(\theta|\mathcal{D}, \omega)$, following data assimilation $\mathcal{D}$, is proportional to the product of the prior weight distribution $p(\theta|\omega)$ and the likelihood $p(\mathcal{D}|\theta)$. Instead of approximation methods such as MCMC~\cite{rasmussen1995practical}, a particular weight prior is chosen corresponding to the weight decay $\omega$. Typically, weight decay is used to regularize DNNs, corresponding to explicit L2 regularization during the SGD~\cite{loshchilov2016sgdr} training process without momentum. When SGD is used with momentum, implicit regularization still exists, producing a vague prior on the weights of the deep neural network. This regularizer can be given an explicit Gaussian-like form~\cite{DBLP:journals/corr/abs-1711-05101}, corresponding to a prior distribution on the weights.


Furthermore, we are sampling the weights $\theta_l$ from the $\epsilon$-likelihood region of the estimated class-conditional distribution:


\begin{equation} 
    \label{eq:virtual}
    \theta_l \sim \frac{1}{(2 \pi)^{m / 2}|\hat{\boldsymbol\Sigma}|^{1 / 2}} e^{\left(-\frac{1}{2}(\theta_{l}-\hat{\bm\mu}_\theta)^{\top} \hat{\boldsymbol\Sigma}^{-1}(\theta_{l}-\hat{\bm\mu}_\theta)\right)} < \epsilon
\end{equation}
\noindent where $\hat{\theta_l} \sim \mathcal{N}(\hat{\bm\mu}_\theta,\hat{\boldsymbol\Sigma})$ denotes the proposed posterior distribution for layer $l$, which are in the sub-level set based on the likelihood. The hyperparameter $\epsilon$ is chosen to be sufficiently small as the sampled weight parameters will not be too far away from the original value causing the potential decrease of the performance of the model.

\begin{table*}[!tpb] \centering    

\begin{tabular}{l|lll|lll}
\toprule
\textbf{}                                                                       & \multicolumn{3}{l|}{\textbf{ID: Berkeley DeepDrive-100k}}                                                                                                      & \multicolumn{3}{l}{\textbf{ID: PASCAL-VOC}}                                                                                                                    \\
\textbf{Method}                                                                 & \multicolumn{1}{l|}{\textbf{FPR95$\downarrow$}}                      & \multicolumn{1}{l|}{\textbf{AUROC$\uparrow$}}                        & \textbf{mAP(ID)} & \multicolumn{1}{l|}{\textbf{FPR95$\downarrow$}}                      & \multicolumn{1}{l|}{\textbf{AUROC$\uparrow$}}                        & \textbf{mAP(ID)} \\  \midrule
Faster-RCNN (Baseline)~\cite{DBLP:journals/pami/RenHG017} & \multicolumn{1}{l|}{82.34}                                           & \multicolumn{1}{l|}{51.39}                                           & 31.2             & \multicolumn{1}{l|}{70.99}                                           & \multicolumn{1}{l|}{58.37}                                           & 48.7             \\
MC Dropout~\cite{Gal2016Uncertainty}                      & \multicolumn{1}{l|}{78.64}                                           & \multicolumn{1}{l|}{53.12}                                           & 31.1             & \multicolumn{1}{l|}{72.19}                                           & \multicolumn{1}{l|}{57.14}                                           & 48.7             \\
SWAG~\cite{maddox2019simple}                                              & \multicolumn{1}{l|}{68.75}                                           & \multicolumn{1}{l|}{57.41}                                           & 31.2             & \multicolumn{1}{l|}{65.14}                                           & \multicolumn{1}{l|}{62.45}                                           & 48.7             \\
BayesOD~\cite{harakeh2019bayesod}                         & \multicolumn{1}{l|}{72.64}                                           & \multicolumn{1}{l|}{55.84}                                           & \textbf{32.4}             & \multicolumn{1}{l|}{60.16}                                           & \multicolumn{1}{l|}{64.62}                                           & \textbf{48.9}             \\
Energy score~\cite{liu2020energy}                         & \multicolumn{1}{l|}{60.06}                                           & \multicolumn{1}{l|}{77.48}                                           & 31.2             & \multicolumn{1}{l|}{56.89}                                           & \multicolumn{1}{l|}{83.69}                                           & 48.7             \\
VOS~\cite{DBLP:journals/corr/abs-2103-02603}              & \multicolumn{1}{l|}{44.27}                                           & \multicolumn{1}{l|}{86.87}                                           & 31.3             & \multicolumn{1}{l|}{47.53}                                           & \multicolumn{1}{l|}{88.7}                                            & \textbf{48.9}             \\
\rowcolor[HTML]{EFEFEF} 
\textbf{Faster-RCNN with Bayesian Layers}                                       & \multicolumn{1}{l|}{\cellcolor[HTML]{EFEFEF}73.42$\pm$2.31}          & \multicolumn{1}{l|}{\cellcolor[HTML]{EFEFEF}56.42$\pm$1.25}          & 31.2             & \multicolumn{1}{l|}{\cellcolor[HTML]{EFEFEF}66.15$\pm$2.15}          & \multicolumn{1}{l|}{\cellcolor[HTML]{EFEFEF}68.63$\pm$2.45}          & 48.7             \\
\rowcolor[HTML]{EFEFEF} 
\textbf{BayesOD with Bayesian Layers}                                           & \multicolumn{1}{l|}{\cellcolor[HTML]{EFEFEF}64.15$\pm$1.69}          & \multicolumn{1}{l|}{\cellcolor[HTML]{EFEFEF}59.26$\pm$0.93}          & \textbf{32.4}             & \multicolumn{1}{l|}{\cellcolor[HTML]{EFEFEF}58.14$\pm$2.31}          & \multicolumn{1}{l|}{\cellcolor[HTML]{EFEFEF}72.12$\pm$1.63}          & \textbf{48.9}             \\
\rowcolor[HTML]{EFEFEF} 
\textbf{Energy Score with Bayesian Layers}                                      & \multicolumn{1}{l|}{\cellcolor[HTML]{EFEFEF}56.42$\pm$2.08}          & \multicolumn{1}{l|}{\cellcolor[HTML]{EFEFEF}82.13$\pm$1.18}          & 31.2             & \multicolumn{1}{l|}{\cellcolor[HTML]{EFEFEF}54.8$\pm$1.25}           & \multicolumn{1}{l|}{\cellcolor[HTML]{EFEFEF}83.99$\pm$0.81}          & 48.7             \\
\rowcolor[HTML]{EFEFEF} 
\textbf{VOS with Bayesian Layers}                                               & \multicolumn{1}{l|}{\cellcolor[HTML]{EFEFEF}\textbf{43.45$\pm$2.21}} & \multicolumn{1}{l|}{\cellcolor[HTML]{EFEFEF}\textbf{87.25$\pm$1.08}} & 31.3    & \multicolumn{1}{l|}{\cellcolor[HTML]{EFEFEF}\textbf{46.55$\pm$1.11}} & \multicolumn{1}{l|}{\cellcolor[HTML]{EFEFEF}\textbf{88.75$\pm$0.60}} & \textbf{48.9}    \\ 
\bottomrule
\end{tabular}

    \caption{\small Main results of the comparison between the proposed method and competitive out-of-distribution detection methods. All methods are implemented based on Faster-RCNN with ResNet-50 as the backbone, and then evaluated on COCO2017 as the OOD dataset. $\uparrow$ indicates larger values are better and $\downarrow$ indicates smaller values are better. All values are percentages. \textbf{Bold} numbers are superior results. We report standard deviations estimated across 5 runs.}
    \label{tab:main_results}
\end{table*}

\subsection{OOD Scoring Rule}
An appropriate scoring rule $S(p_{\boldsymbol\theta} (\mathbf{x},\mathbf{b}))$ is required to distinguish the OOD from the ID data. Scoring rules can be further divided into \textit{local} and \textit{non-local} rules~\cite{harakeh2021estimating}. As a local scoring rule, the negative Log Likelihood (NLL) function~\cite{lakshminarayanan2017simple} evaluates a predictive distribution based on its value only at the true target. NLL measure the quality of predicted probability distributions of a test dataset by: $\text{NLL}=-\frac{1}{N_{test}}\sum_{n=1}^{N_{test}} \log p(\mathbf{y}_n|\mathbf{x}_n, \mathcal{D})$, where $\mathbf{x}_n$ is a test data point, and $\mathbf{y}_n$ its corresponding ground truth label. NLL ranges in $(-\infty, +\infty)$, with a lower NLL score indicating a better fitting predictive distribution for that specific ground truth label. 
However, NLL has its limitation in learning and evaluating bounding box predictive distributions. Energy Score~\cite{gneiting2007strictly} is a proper and non-local scoring rule as an alternative for learning and evaluating multivariate Gaussian predictive distributions. Energy score uses a non-probabilistic energy function to attribute lower values to in-distribution data and higher values to out-of-distribution data. Energy score has been proven efficient compared to the standard softmax score for OOD detection on both image classification and object detection tasks~\cite{liu2020energy, DBLP:journals/corr/abs-2103-02603}. For object detection tasks, the object-level energy score function~\cite{DBLP:conf/nips/LiuWOL20} can be defined as
\begin{equation}
    E(\mathbf{x},\mathbf{b};\theta) = -T \log \sum_{k=1}^K {\exp(f_k((\mathbf{x},\mathbf{b});\theta))}
    \label{eq:es}
\end{equation}
where the $f_k((\mathbf{x},\mathbf{b});\boldsymbol\theta)$ is the logit output for class $k$ in the classification head. $T$ represents the temperature parameter of the Gibbs distribution~\cite{liu2020energy}. In this case, it is set as a constant hyperparameter.
During inference, given a test input $\hat{\mathbf{x}}$, the object detector produces a bounding box prediction $\hat{\mathbf{b}}$ through the Bayesian model consists of sampled parameters $\theta_l$. The OOD uncertainty score for the predicted object $(\*x^*, \*b^*)$ is given by:
\begin{align}
    S(p_\theta(\hat{\mathbf{x}}), \hat{\mathbf{b}}) = \frac{\exp (- \phi (E(\hat{\mathbf{x}},\hat{\mathbf{b}};{\theta_l})))}{1+\exp(-  \phi(E(\hat{\mathbf{x}},\hat{\mathbf{b}};{\theta_l})))}.
    \label{eq:ood_uncertainty}
    \vspace{-0.5em}
\end{align}
$\phi(\cdot)$ can be a trainable nonlinear function that allows flexible energy surface learning. In this case, $\phi(\cdot)$ is set to be a hyperparameter for inference only.

For OOD detection, ID and OOD objects are distinguished through the following equation:
\begin{equation}
    G(\hat{\mathbf{x}},\hat{\mathbf{b}})=\left\{\begin{array}{ll}
1 & \text { if }S(p_\theta(\hat{\mathbf{x}}), \hat{\mathbf{b}})\geq \gamma, \\
0 & \text { if }S(p_\theta(\hat{\mathbf{x}}), \hat{\mathbf{b}}) <\gamma.
\end{array}\right.
\label{eq:ood_detection}
\end{equation}
where $G(\hat{\mathbf{x}},\hat{\mathbf{b}})=1$ implies that the current object is ID, and $G(\hat{\mathbf{x}},\hat{\mathbf{b}})=0$ means the OOD data. The threshold $\gamma$ is typically chosen so that a high fraction of ID data (e.g., 95\%) is correctly classified.

\begin{figure*}[t]
	\centering
        \includegraphics[width=0.8\textwidth, height = 5.2cm]{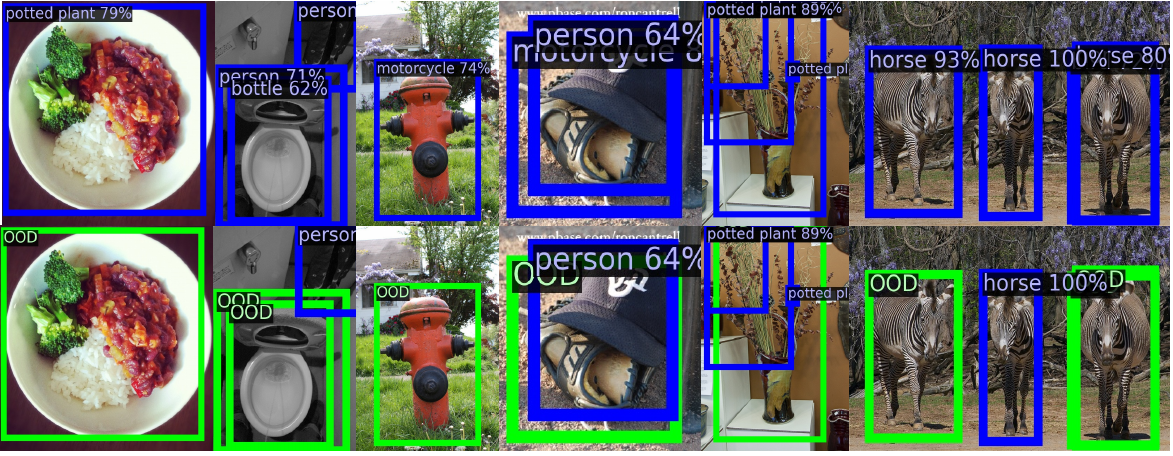}
	\caption{The visualization results of OOD detection. Images are taken from COCO2017. The first row of visualization results is provided by a vanilla Faster-RCNN. The bottom row of visualization results is produced when inference with Bayesian Layers. Models are trained on PASCAL-VOC and evaluated on COCO. Detected objects with blue bounding boxes are classified as one of the ID classes, while objects detected in green bounding boxes indicate that they are recognized as OOD objects.} 
        \label{fig:ood}
\end{figure*}

\section{Experimental Results}\label{sec:results}
In this section, we demonstrate the effectiveness of our proposed Bayesian inference method on a wide range of OOD benchmarks with different base models for both object detection and image classification tasks, respectively in Section~\ref{sec:OD} and in Section~\ref{sec:IC}. The visualization results are shown in Figure~\ref{fig:ood}.

\subsection{Evaluation on Object Detection}\label{sec:OD}
\paragraph{Experimental Setup}
To evaluate our method, we use \texttt{PASCAL VOC}\footnotemark[1]~\cite{DBLP:journals/ijcv/EveringhamGWWZ10} and \texttt{Berkeley DeepDrive 100K (BDD-100k\footnotemark[2])}~\cite{DBLP:conf/cvpr/YuCWXCLMD20} datasets as the ID training data, \texttt{MS-COCO}~\cite{lin2014microsoft} as the OOD data. PASCAL VOC 2007 and PASCAL VOC 2012 containing 21493 images for train/validation/test data are used. BDD-100k contains 70,000/10,000/20,000 images for train/val/test with 1.8M objects. COCO2017 includes 123,287 images and 886,284 instances.

The models are implemented based on the Detectron2 library~\cite{Detectron2018}. Faster-RCNN with ResNet-50 backbone~\cite{he2016identity} is used as the base framework for object detection. We choose sample numbers of 1000 for each single test run. The Bayesian layers $W_l$ are chosen to be the 2D Convolutional layers in the main results listed in Table~\ref{tab:main_results}. Faster R-CNN model with a ResNet-50 backbone has a total of 52 Conv2D layers including 49 Conv2D layers from the ResNet-50 backbone and 3 Conv2D layers from the RPN and detection head. The impact of selecting different layers to be Bayesian is discussed in Section~\ref{sec:layers}.
\paragraph{Evaluation Metrics} 
Two evaluation metrics are used to evaluate the performance of proposed models on OOD detection, which are: (1)FPR95, the false positive rate of OOD samples when the true positive rate of ID samples is at 95\%; (2)AUROC, computes Area Under the Receiver Operating Characteristic Curve. In addition, mean average precision (mAP) is reported to represent the performance on the ID task.

\footnotetext[1]{\color{black}PASCAL-VOC consists of the following ID labels: {Person, Car, Bicycle, Boat, Bus, Motorbike, Train, Airplane, Chair, Bottle, Dining Table, Potted Plant, TV, Sofa, Bird, Cat, Cow, Dog, ~Horse, Sheep}.}
\footnotetext[2]{\color{black}BDD-100k consists of ID labels: {Pedestrian, Rider, Car, Truck, Bus, Train, Motorcycle, Bicycle, Traffic light, Traffic sign}.}

\paragraph{Bayesian Layers are Effective on OOD Detection} 
Table~\ref{tab:main_results} shows that in comparison to different pre-trained base models, algorithms with Bayesian layers can provide varying degrees of performance enhancement. Our proposed method is compared with several baselines introduced in Section~\ref{sec:bg}, including Faster-RCNN (Baseline)~\cite{DBLP:journals/pami/RenHG017}, MC Dropout~\cite{Gal2016Uncertainty}, SWAG~\cite{maddox2019simple}, BayesOD~\cite{harakeh2019bayesod}, Energy score~\cite{liu2020energy}, and VOS~\cite{DBLP:journals/corr/abs-2103-02603}. Several popular Bayesian deep learning methods often used as benchmarks are included, as well as outperforming OOD detection algorithms as comparisons to prove the efficiency of inference with Bayesian layers on OOD detection. The comparison precisely highlights the benefits of incorporating Bayesian layers for OOD inference. The experimental results show that all baselines inferred by Bayesian layers have increased OOD detection performance.  Results shown in Table~\ref{tab:cls_results} are produced based on the convolutional layers from the backbone and the detection head chosen to be transformed into Bayesian layers. 

Compared to each base model, inference with Bayesian layers increases the performance for both experiments with two ID datasets. Some baselines rely on a classification model trained primarily for the ID classification task. Due to the existence of a classification head, these methods can be naturally extended to the object detection model. MC Dropout and SWAG are similar Bayesian methods to our method which is also scalable to deep learning models and focus on the posterior distribution approximation during the inference stage. Compared to MC Dropout, inference with Bayesian Layers on Faster-RCNN improved the FPR95 scores by \textbf{7.53}\% on BDD-100k and \textbf{8.19}\% on Pascal VOC, and AUROC scores by \textbf{4.55}\% on BDD-100k and \textbf{13.94}\% on Pascal VOC. Moreover, our method preserves the high accuracy on the original in-distribution task (measured by mAP) as long as choosing the appropriate baseline.

Most existing OOD detection methods focus on modeling a single type of uncertainty. The epistemic uncertainty comes from the classification branch modeling the classifier or the regression branch modeling the bounding box regression. Other methods, such as VOS, rely on synthesis data outliers to measure the aleatoric uncertainty. Inference with Bayesian layers makes it possible to quantify the uncertainties from the classification head and the regression head at the same time by choosing to transform Bayesian Layers from backbone and detection heads together. This provides a theoretical explanation of the performance improvement of Bayesian layers being applied to all baselines. Inference with Bayesian layers offers a tool to model the epistemic uncertainty coming out of every corner of the network architecture. Combined with the aleatoric uncertainties modeling allows further performance improvement for OOD detection.

\paragraph{Impact of Bayesian Layers Selection}\label{sec:layers}
\begin{table}[!tpb]
\centering
\renewcommand\arraystretch{1.3}
\scalebox{0.9}
 {
\begin{tabular}{l|c}
\toprule
\textbf{Inference Layer Selection}                        & \textbf{FPR95$\downarrow$ / AUROC$\uparrow$} \\ \midrule
Deterministic Faster-RCNN (Baseline)                      & 82.34 / 51.39          \\
Conv2D layers (Backbone) as Bayesian Layers               & 73.73 / 56.14          \\
Linear layers (Backbone) as Bayesian Layers               & 75.25 / 54.73          \\
\textbf{Conv2D layers (Backbone+head) as Bayesian Layers} & \textbf{73.42 / 56.42} \\
Linear layers (Backbone+head) as Bayesian Layers          & 75.82 / 54.12          \\
Full layers as Bayesian Layers                            & 74.12 / 55.73          \\ 
\bottomrule
\end{tabular}
}
\caption{\small FPR and AUROC results of different inference Bayesian layers selections. BDD 100k is the ID training data, COCO is the OOD data.}
\label{tab:layer_results}
\end{table}

To further investigate the impact of choosing different layers to be transformed into Bayesian layers during the Bayesian inference, we test five different architectures on COCO as the OOD dataset when trained on BDD-100k as ID data. Table~\ref{tab:layer_results} shows the mean FPR95 and AUROC results of 3 runs. Specifically, we consider two types of neural network layers: (i) Convolutional Layer: As the primary building block of a CNN. The convolutional layer computes the convolutional operation of the input images using kernel filters to extract fundamental features. The model captures epistemic uncertainties when meeting the OOD data by replacing deterministic weight parameters with the proposed Gaussian distributions. (ii) Linear Layer: A linear layer, also known as a fully connected layer. It is capable of learning an average rate of correlation between the output and the input without a bias. Linear layers are frequently modified by Bayesian methods since they contain limited parameters leading to less training and computational inference cost. 
\begin{table*}[!tbp]
    \centering    
\begin{tabular}{llll}
\toprule
                                                                                                                         &                                                                     & \textbf{FPR95}                                & \textbf{AUROC}            \\
\multirow{-2}{*}{\textbf{\begin{tabular}[c]{@{}l@{}}Base model \\ $\mathcal{D}_{\text{in}}^{\text{test}}$\end{tabular}}} & \multirow{-2}{*}{\textbf{Method}}                                   & \cellcolor[HTML]{FFFFFF}\textbf{$\downarrow$} & $\uparrow$                \\ \midrule
                                                                                                                         & \multicolumn{1}{l|}{WideResNet (Baseline)}                          & 51.04                                         & 90.90                     \\
                                                                                                                         & \multicolumn{1}{l|}{Bayesian Neural Networks}                       & 53.04                                         & 92.45                     \\
                                                                                                                         & \multicolumn{1}{l|}{MC dropout}                                     & 54.32                                         & 88.90                     \\
                                                                                                                         & \multicolumn{1}{l|}{SWAG}                                           & 38.25                                         & 92.91                     \\
                                                                                                                         & \multicolumn{1}{l|}{Energy score}                                   & 33.01                                         & 91.88                     \\
                                                                                                                         & \multicolumn{1}{l|}{Vitual Outliers}                                & 24.87                                         & 94.06                     \\
                                                                                                                         & \multicolumn{1}{l|}{\textbf{Baseline + Bayesian Layers (Ours)}}     & \textbf{48.25  $\pm$ 0.91}                    & \textbf{92.27 $\pm$ 2.24} \\
                                                                                                                         & \multicolumn{1}{l|}{\textbf{Energy Score + Bayesian Layers (Ours)}} & \textbf{32.33  $\pm$ 0.46}                    & \textbf{93.52 $\pm$ 1.82} \\
\multirow{-9}{*}{\begin{tabular}[c]{@{}l@{}}\textbf{WideResNet}\\ \\ CIFAR-10\end{tabular}}                                       & \multicolumn{1}{l|}{\textbf{VOS + Bayesian Layers (Ours)}}          & \textbf{23.68 $\pm$ 0.52}                     & \textbf{94.36 $\pm$ 0.93} \\ 
\bottomrule
\end{tabular}

    \caption{\small OOD detection results of the proposed Bayesian methods on image classification. All models listed are built based on the WideResNet-40 architecture. All models are first trained on CIFAR-10 dataset and then evaluated on several OOD datasets. The mean PR95 and AUROC results are given.}
    \label{tab:cls_results}
\end{table*}

\subsection{Evaluation on Image Classification}\label{sec:IC}
We demonstrate that Bayesian Layers are also effective at common image classification benchmarks beyond object detection. In this case, CIFAR-10~\cite{cifar} with standard train/val splits is used as the ID training data. WideResNet-40~\cite{zagoruyko2016wide} is chosen to be the base neural network architecture. For image classification, the cross-entropy loss is used as the training objective. All models are evaluated on six OOD datasets:\texttt{Textures}~\cite{DBLP:conf/cvpr/CimpoiMKMV14}, \texttt{SVHN}~\cite{netzer2011reading}, \texttt{Places365}~\cite{DBLP:journals/pami/ZhouLKO018}, \texttt{LSUN-C}~\cite{DBLP:journals/corr/YuZSSX15}, \texttt{LSUN-Resize}~\cite{DBLP:journals/corr/YuZSSX15}, and
\texttt{iSUN}~\cite{DBLP:journals/corr/XuEZFKX15}. The comparisons are shown in Table~\ref{tab:cls_results}, with results averaged over six test datasets. We demonstrate that inferencing with Bayesian layers for OOD detection has different levels of performance improvement compared to each benchmark, without sacrificing the ID test classification accuracy (94.84\% on pre-trained WideResNet) and extra training requirement.
\section{Conclusion}\label{sec:conclusion}

In this work, we propose a novel and scalable Bayesian deep learning inference framework for OOD detection. Transforming the deterministic deep neural layers into Bayesian layers by replacing the weights parameters trained by point estimates with assumed Gaussian distributions is able to provide the uncertainty estimation during the inference stage. We show that inference with superficial Bayesian layers can improve the performance of various benchmarks for OOD detection. Unlike other Bayesian deep learning methods, our proposed method does not require extra training, making it scalable for modern advanced deep-learning-based object detectors. For future work, we would like to explore the impact on different assumed distributions for Bayesian layers and the uncertainty quantification method for Bayesian layers. In addition, our approach can be valuable to other machine learning tasks than object detection or image classification. We hope future research will increase attention toward a broader view of OOD uncertainty estimation from a Bayesian layers perspective.


\textbf{Acknowledgements} We are grateful to the UK EPSRC Council via the  project EP/V026747/1 (Trustworthy Autonomous Systems Node in Resilience. We are also grateful to the UK EPSRC for
funding this work through the EP/T013265/1 project
NSF-EPSRC: “ShiRAS. Towards Safe and Reliable
Autonomy in Sensor Driven” and the support for
ShiRAS by the National Science Foundation under
Grant USA NSF ECCS 1903466. For the purpose
of open access, the authors have applied a Creative
Commons Attribution (CC BY) licence to any Author
Accepted Manuscript version arising.


\bibliographystyle{IEEEtran}
\bibliography{10_bibliography}


\end{document}